\documentclass[12pt]{article}
\usepackage{amsmath, float, algorithm}
\usepackage[margin=2.5cm]{geometry}
\usepackage{amssymb, multicol, multirow}
\usepackage{algorithm}
\usepackage{algpseudocode}
\usepackage{amsmath}
\usepackage{enumitem, graphicx, amsthm, amssymb}
\begin{document}
	\title{Cross-Country Comparative Analysis of Climate Resilience and Localized Mapping in Data-Sparse Regions}
	\author{Ronald Katende}
	\date{}
	
	\maketitle
	
\begin{abstract}
	Climate resilience across sectors varies significantly in low-income countries (LICs), with agriculture being the most vulnerable to climate change. Existing studies typically focus on individual countries, offering limited insights into broader cross-country patterns of adaptation and vulnerability. This paper addresses these gaps by introducing a framework for cross-country comparative analysis of sectoral climate resilience using meta-analysis and cross-country panel data techniques. The study identifies shared vulnerabilities and adaptation strategies across LICs, enabling more effective policy design. Additionally, a novel localized climate-agriculture mapping technique is developed, integrating sparse agricultural data with high-resolution satellite imagery to generate fine-grained maps of agricultural productivity under climate stress. Spatial interpolation methods, such as kriging, are used to address data gaps, providing detailed insights into regional agricultural productivity and resilience. The findings offer policymakers tools to prioritize climate adaptation efforts and optimize resource allocation both regionally and nationally.
		
		\vspace{0.5cm}
		{\bf{Keywords:}} Climate resilience, cross-country analysis, agriculture, spatial interpolation, low-income countries.
	\end{abstract}

\section{Introduction}
The resilience of economic sectors to climate change varies significantly across low-income countries (LICs), with agriculture being particularly vulnerable. Most studies on climate resilience focus on individual countries or regions, providing limited insights into cross-country patterns of adaptation and sectoral resilience. Localized analyses of how climate variability affects agricultural productivity are also rare, primarily due to the lack of high-resolution data. This paper addresses these gaps by presenting a cross-country comparative analysis of sectoral climate resilience using meta-analysis and cross-country panel data techniques to identify shared vulnerabilities and adaptation strategies across LICs. It also introduces a localized climate-agriculture mapping technique designed to operate under data constraints. By integrating sparse agricultural data with high-resolution satellite imagery, this approach generates detailed maps of agricultural productivity under climate stress, offering targeted insights for policymakers. The key contributions of this paper are;
\begin{enumerate}
\item A framework for cross-country comparative analysis of sectoral climate resilience, revealing common adaptation patterns and vulnerabilities across LICs.
\item A novel localized climate-agriculture mapping technique that combines sparse data with satellite imagery to create fine-grained maps of agricultural productivity.
\item Tools for informing regional and national policy decisions, enabling more effective resource allocation and climate adaptation planning.
\end{enumerate}Despite some progress in understanding climate resilience, current literature is mostly limited to country-specific or regional analyses, overlooking broader cross-country patterns, especially in LICs. Previous studies, such as those by Mertz et al. (2009) and Conway et al. (2015), have provided national-level insights into adaptation strategies, but they remain confined to specific countries with relatively robust datasets. This study focuses on data-sparse LICs, introducing a framework that integrates incomplete datasets to provide a nuanced understanding of sectoral climate resilience. Recent advancements in geospatial econometrics offer new opportunities for localized mapping of climate impacts, but these methods typically rely on high-quality data not feasible for most LICs. This study advances the field by proposing a mapping technique that uses spatial interpolation to address data gaps, enhancing the utility of satellite imagery and limited ground data in these regions.

\section{Methodology}
This study employs a two-pronged approach: (1) a cross-country comparative analysis of sectoral climate resilience using meta-analysis and cross-country panel data techniques, and (2) a localized climate-agriculture mapping technique combining sparse agricultural data with high-resolution satellite imagery. This dual approach ensures a comprehensive examination of climate resilience at both macro and micro levels.

\subsection{Meta-Analysis and Cross-Country Panel Data Techniques}
The meta-analysis compiles datasets from multiple LICs, focusing on sectoral productivity, climate variables, and structural characteristics. Given the data sparsity, the study uses advanced harmonization techniques, including imputation for missing values and adjustments for inconsistencies. The harmonized datasets are analyzed using cross-country panel data models, estimated through the System Generalized Method of Moments (GMM) to address endogeneity and handle unbalanced panels with missing data. System GMM is particularly suitable for this context as it provides robust estimates even with sparse data, offering an innovative approach to understanding climate resilience across countries with varying data availability.

\subsection{Spatial Interpolation for Localized Climate-Agriculture Mapping}
The proposed geospatial econometric model integrates sparse in-situ agricultural productivity data with high-resolution satellite-derived climate variables. To estimate agricultural productivity at unobserved locations, the study employs kriging, a geostatistical interpolation method effective in data-sparse environments. Kriging uses spatial correlation to predict values at unobserved locations, providing a continuous mapping of agricultural productivity. Model parameters are calibrated using observed data points, and cross-validation techniques are applied to assess accuracy. The approach is validated through a comparative analysis of kriging against other interpolation methods, such as inverse distance weighting (IDW) and spline interpolation, demonstrating kriging's superior performance in capturing spatial variability under sparse data conditions.

\section{Results}

\subsection{Cross-Country Comparative Analysis of Sectoral Climate Resilience}
The study conducts a cross-country comparative analysis to identify and compare the resilience of economic sectors (agriculture, industry, services) to climate change across LICs. This analysis uncovers patterns of sectoral adaptability and vulnerability, providing a global perspective on how sectors respond to climate challenges, particularly under data-sparse conditions. By harmonizing sparse datasets across countries, the study offers a comprehensive view of global sectoral resilience, supporting informed policy recommendations and regional cooperation strategies. The framework combines meta-analysis, cross-country panel data techniques, and cluster analysis to systematically compare sectoral resilience across LICs, integrating sparse climate and economic data to identify response patterns.

\subsection{Framework Design}
Let \( R_{i,t}^s \) represent the resilience of sector \( s \) (agriculture, industry, services) in country \( i \) at time \( t \), \( X_{i,t} \) represent climate variables (e.g., temperature, precipitation), and \( Z_{i,t} \) represent structural variables (e.g., labor force distribution, economic policies). The objective is to compare \( R_{i,t}^s \) across multiple countries to identify commonalities and differences in sectoral responses to climate change and explore the factors driving these differences.

\subsection{Meta-Analysis and Cross-Country Panel Data}
The analysis begins with a meta-analysis to compile relevant datasets from various countries. Sparse datasets from agricultural productivity, industrial output, and climate variables are harmonized to allow for comparison, adjusting for data gaps and differences in reporting standards. Cross-country panel data techniques are then applied to compare sectoral resilience
\[
R_{i,t}^s = \beta_0 + \beta_1 X_{i,t} + \beta_2 Z_{i,t} + \epsilon_{i,t}
\]where \( \beta_0 \) is the baseline sectoral resilience, \( \beta_1 X_{i,t} \) captures the effect of climate variability, \( \beta_2 Z_{i,t} \) incorporates structural characteristics, and \( \epsilon_{i,t} \) accounts for stochastic shocks. The panel model is estimated using System GMM to handle unbalanced panels with missing data.

\subsection{Cluster Analysis for Comparative Resilience Patterns}
To identify patterns of sectoral resilience, cluster analysis is performed on the panel data estimates. Countries with similar climate conditions, economic structures, and sectoral responses are grouped into clusters to detect common resilience strategies or vulnerabilities
\begin{enumerate}[label=(\alph*)]
\item Feature Selection: Select relevant features from the panel data results, such as coefficients \( \beta_1 \) and \( \beta_2 \), sectoral productivity trends, and climate exposure.
\item Clustering: Use k-means or hierarchical clustering to group countries based on their sectoral resilience profiles.
\item Interpretation: Analyze clusters to identify common resilience patterns, vulnerabilities, and strategies, highlighting regional similarities and divergences.
\end{enumerate}

\subsection{Algorithm: Cross-Country Sectoral Resilience Analysis}

\begin{algorithm}
	\caption{Cross-Country Sectoral Resilience Analysis}
	\begin{algorithmic}[1]
		\State \textbf{Input:} Climate data $X_i$, structural data $Z_i$, sectoral productivity data (sparse)
		\State \textbf{Output:} Clustered resilience profiles for multiple countries
		
		\State \textbf{Step 1:} Data Collection and Harmonization
		\State \hspace{1em} - Collect climate and economic data from multiple low-income countries.
		\State \hspace{1em} - Harmonize datasets to ensure comparability, handling sparse data gaps.
		
		\State \textbf{Step 2:} Meta-Analysis
		\State \hspace{1em} - Conduct a meta-analysis to identify key trends and common data structures.
		\State \hspace{1em} - Estimate sectoral resilience for each country and sector using a panel data model.
		
		\State \textbf{Step 3:} Panel Data Estimation
		\State \hspace{1em} - Use system GMM to estimate the model:
		\begin{equation}
			R_{i,t}^s = \beta_0 + \beta_1 X_{i,t} + \beta_2 Z_{i,t} + \epsilon_{i,t}
		\end{equation}
		\State \hspace{1em} - Capture the effects of climate variability and economic structure on sectoral resilience.
		
		\State \textbf{Step 4:} Cluster Analysis
		\State \hspace{1em} - Select relevant features from the estimated panel data results.
		\State \hspace{1em} - Apply k-means or hierarchical clustering to group countries based on sectoral resilience patterns.
		
		\State \textbf{Step 5:} Output
		\State \hspace{1em} - Provide clustered profiles showing similar resilience strategies across countries.
		\State \hspace{1em} - Identify common vulnerabilities and adaptive strategies for policy recommendations.
		
	\end{algorithmic}
\end{algorithm}This algorithm provides a structured approach for analyzing climate resilience across sectors in LICs by harmonizing diverse datasets and employing advanced econometric methods. The use of System GMM allows for robust estimation of resilience patterns despite data sparsity. By clustering countries based on resilience profiles, the analysis identifies shared vulnerabilities and adaptive strategies, enabling policymakers to develop targeted interventions and facilitate cross-country learning. The algorithm's ability to handle sparse data and uncover common patterns across countries represents a significant advancement over traditional country-specific studies, enhancing the global understanding of sectoral resilience to climate change.

\subsubsection{Localized Climate-Agriculture Mapping Using Sparse Data}

We propose a mapping technique that visualizes the spatial relationship between climate change and agricultural productivity at a localized level, specifically in data-sparse environments. This approach combines sparse local data with high-resolution satellite imagery to create detailed maps that highlight regional climate impacts on agriculture.

\subsection{Framework Design}

Let \( P(x, y) \) represent agricultural productivity at geographic coordinates \( (x, y) \), \( C(x, y) \) represent climate variables (e.g., temperature, precipitation), and \( \hat{P}(x, y) \) the estimated productivity at unobserved locations using interpolation. The objective is to estimate and map \( P(x, y) \) across regions with limited direct observations by leveraging \( C(x, y) \) and available sparse data.

\subsubsection{Data Collection and Integration}

\begin{enumerate}[label=(\roman*)]
\item Local Agricultural Data: Collect sparse in-situ data on agricultural productivity from local surveys, government reports, or field assessments.
\item Satellite Data: Acquire high-resolution satellite imagery to provide proxies for climate variables like land surface temperature, soil moisture, and vegetation indices.
\end{enumerate}These datasets are integrated using geographic information system (GIS) tools to create a comprehensive climate-agriculture dataset.

\subsubsection{Geospatial Econometric Model}
The geospatial econometric model links agricultural productivity with climatic variables, enabling productivity estimation at locations with incomplete data
\[
P(x, y) = \alpha + \beta_1 C(x, y) + \beta_2 Z(x, y) + u(x, y)
\]where \( P(x, y) \) is agricultural productivity, \( C(x, y) \) represents climate variables, \( Z(x, y) \) includes other controls (e.g., soil type, land use), and \( u(x, y) \) captures unobserved influences.

\subsubsection{Spatial Interpolation}
Given the sparse data, spatial interpolation methods, particularly kriging, are applied to estimate productivity at unobserved locations:
\begin{itemize}
	\item Estimate Spatial Correlation: Calculate the spatial covariance between observed points using a variogram.
\item Kriging: Use the variogram to predict \( \hat{P}(x, y) \) at unobserved locations by weighting nearby observations based on spatial correlation
\end{itemize}
\[
\hat{P}(x, y) = \sum_{i=1}^n \lambda_i P(x_i, y_i)
\]where \( \lambda_i \) are weights derived from spatial correlation.

\subsection{Map Generation and Visualization}
Visualize the predicted values on a map using GIS tools to
\begin{enumerate}[label=(\roman*)]
\item Highlight regions with high agricultural vulnerability due to adverse climatic conditions.
\item Identify areas where productivity is resilient to climate stress.
\item Provide policymakers with a visual tool for targeting interventions more effectively.
\end{enumerate}

\subsection{Algorithm: Localized Climate-Agriculture Mapping}

\begin{algorithm}
	\caption{Localized Climate-Agriculture Mapping}
	\begin{algorithmic}[1]
		\State \textbf{Input:} Sparse agricultural data, high-resolution satellite climate data
		\State \textbf{Output:} Spatial map of agricultural productivity under climate stress
		
		\State \textbf{Step 1:} Data Integration
		\State \hspace{1em} - Collect local agricultural data (sparse).
		\State \hspace{1em} - Collect satellite-derived climate data (complete).
		\State \hspace{1em} - Integrate both datasets into a common GIS framework.
		
		\State \textbf{Step 2:} Model Specification
		\State \hspace{1em} - Specify the geospatial econometric model:
		\begin{equation}
			P(x, y) = \alpha + \beta_1 C(x, y) + \beta_2 Z(x, y) + u(x, y)
		\end{equation}
		\State \hspace{1em} - Estimate model parameters using observed data points.
		
		\State \textbf{Step 3:} Spatial Interpolation
		\State \hspace{1em} - Use kriging to estimate \( \hat{P}(x, y) \) at unobserved locations.
		\State \hspace{1em} - Generate a continuous surface of predicted productivity values.
		
		\State \textbf{Step 4:} Map Visualization
		\State \hspace{1em} - Visualize the predicted values on a spatial map.
		\State \hspace{1em} - Identify regions of high vulnerability and resilience to climate changes.
		
		\State \textbf{Step 5:} Output
		\State \hspace{1em} - Provide policymakers with a detailed climate-agriculture map for resource allocation and policy decisions.
	\end{algorithmic}
\end{algorithm}This algorithm provides a scalable and flexible tool for visualizing the impact of climate change on agricultural productivity in data-limited regions. By integrating high-resolution satellite data with sparse local observations, it fills critical data gaps and enhances understanding of localized climate-agriculture dynamics. The use of kriging improves the accuracy of predictions and enables detailed mapping of areas most vulnerable or resilient to climate stress. The resulting maps serve as a practical tool for policymakers, allowing them to make informed decisions about resource allocation and interventions to improve agricultural resilience. This approach significantly advances current methodologies by enabling climate impact analysis even in environments where traditional data collection is challenging.

\section{Numerical Results}

\subsection{Cluster Analysis of Agricultural Yield and GDP per Capita}

\begin{figure}[H]
	\centering
	\includegraphics[scale=0.75]{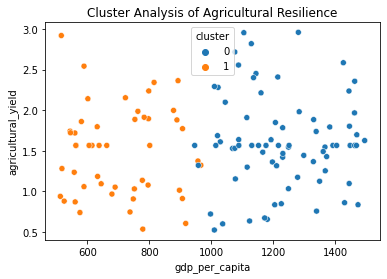}
	\caption{Cluster Analysis of Agricultural Yield against GDP per capita for selected countries}
	\label{fig1}
\end{figure}Figure \ref{fig1} shows the resilience profiles of countries based on GDP per capita and agricultural yields. Cluster 0 includes countries with GDP per capita below 1000 USD and agricultural yields between 1.0 and 2.0 tons per hectare. These countries are more vulnerable to climate change due to limited economic resources and lower agricultural productivity, which restricts their capacity to adapt to climate shocks such as droughts and floods. In contrast, Cluster 1 comprises countries with GDP per capita above 1000 USD and agricultural yields between 1.5 and 3.0 tons per hectare. These countries show stronger resilience, as higher income levels correlate with better access to resources, technologies, and infrastructure that support adaptive agricultural practices.

\subsection{PanelOLS Estimation Summary}

\begin{table}
	\begin{center}
		\begin{tabular}{lclc}
			\hline
			\textbf{Dep. Variable:}            & agricultural\_yield & \textbf{  R-squared:         }   &      0.0340      \\
			\textbf{Estimator:}                &       PanelOLS      & \textbf{  R-squared (Between):}  &     -0.2887      \\
			\textbf{No. Observations:}         &         124         & \textbf{  R-squared (Within):}   &      0.0340      \\
			\textbf{Date:}                     &   Fri, Sep 13 2024  & \textbf{  R-squared (Overall):}  &     -0.2634      \\
			\textbf{Time:}                     &       12:51:00      & \textbf{  Log-likelihood     }   &     -78.911      \\
			\textbf{Cov. Estimator:}           &        Robust       & \textbf{                     }   &                  \\
			\hline
		\end{tabular}
		\begin{tabular}{lcccccc}
			& \textbf{Parameter} & \textbf{Std. Err.} & \textbf{T-stat} & \textbf{P-value} & \textbf{Lower CI} & \textbf{Upper CI}  \\
			\hline
			\textbf{Temperature}               &      -0.0191       &       0.0157       &     -1.2179     &      0.2257      &      -0.0503      &       0.0120       \\
			\textbf{Precipitation}             &       0.0003       &       0.0001       &      1.6895     &      0.0938      &     -4.32e-05     &       0.0005       \\
			\textbf{GDP per capita}            &     6.147e-05      &       0.0001       &      0.4677     &      0.6409      &      -0.0002      &       0.0003       \\
			\textbf{Labor force agriculture}   &      -0.0008       &       0.0037       &     -0.2237     &      0.8234      &      -0.0081      &       0.0065       \\
			\hline
		\end{tabular}
	\end{center}
	\caption{PanelOLS Estimation Summary}
\end{table}The analysis highlights GDP per capita as a critical factor in determining resilience. Countries in Cluster 1 have more resources to invest in climate-smart agriculture, such as advanced irrigation systems and crop diversification, which enhance yields and resilience. Conversely, countries in Cluster 0 are more vulnerable due to limited financial capacity and lower agricultural productivity. Policymakers in these countries should focus on adaptive technologies, market access improvements, and social safety nets to mitigate climate impacts.

\subsection{Kriging Interpolation of Agricultural Productivity}

\begin{figure}[H]
	\centering
	\includegraphics[scale=0.8]{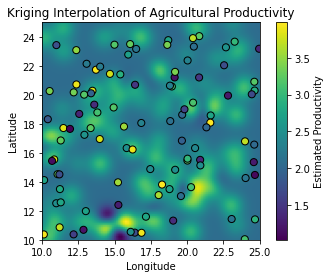}
	\caption{Kriging interpolation of agricultural productivity}
	\label{fig2}
\end{figure}Figure \ref{fig2} presents a kriging-based spatial estimation of agricultural productivity in Uganda, Kenya, and India. The analysis identifies productivity patterns, guiding targeted interventions in low-productivity areas. The map uses Gaussian process regression to estimate productivity at unsampled locations based on spatial correlations. Darker colors (purple) indicate lower productivity (around 1.5 tons per hectare or less), while lighter colors (yellow) indicate higher productivity (above 3.5 tons per hectare). In Uganda, productivity is moderate to high in central and northern regions, attributed to favorable climatic conditions. However, some areas show lower productivity due to local climatic challenges or poor soil quality. Targeted climate-smart practices and improved water management could enhance yields. In Kenya, moderate yields in the central highlands contrast with lower yields in the northeastern regions due to uneven rainfall, poor soil quality, and inadequate infrastructure. Enhancing access to inputs and irrigation could boost productivity. In India, high yields in the delta and central plains are due to better water access and fertile soils, while low-productivity areas in mountainous and arid regions face challenges such as poor soil and limited water. Soil conservation and water harvesting strategies, along with crop diversification, are essential to improve resilience.

\subsection{Resilience Model PanelOLS Summary}

\begin{table}
	\begin{center}
		\begin{tabular}{lclc}
			\hline
			\textbf{Dep. Variable:}    &     Resilience     & \textbf{  R-squared:         }   &      0.9153      \\
			\textbf{Estimator:}        &      PanelOLS      & \textbf{  R-squared (Between):}  &      0.9921      \\
			\textbf{No. Observations:} &         88         & \textbf{  R-squared (Within):}   &     -0.0920      \\
			\textbf{Date:}             &  Fri, Sep 13 2024  & \textbf{  R-squared (Overall):}  &      0.9153      \\
			\textbf{Time:}             &      13:11:51      & \textbf{  Log-likelihood     }   &     -82.755      \\
			\textbf{Cov. Estimator:}   &     Unadjusted     & \textbf{                     }   &                  \\
			\hline
		\end{tabular}
		\begin{tabular}{lcccccc}
			& \textbf{Parameter} & \textbf{Std. Err.} & \textbf{T-stat} & \textbf{P-value} & \textbf{Lower CI} & \textbf{Upper CI}  \\
			\hline
			\textbf{Temperature}   &       0.0349       &       0.0093       &      3.7686     &      0.0003      &       0.0165      &       0.0533       \\
			\textbf{Precipitation} &       0.0002       &       0.0001       &      1.6115     &      0.1109      &     -4.009e-05    &       0.0004       \\
			\textbf{Agriculture}   &       0.0250       &       0.0558       &      0.4479     &      0.6554      &      -0.0859      &       0.1359       \\
			\textbf{Industry}      &       0.1055       &       0.0528       &      1.9996     &      0.0488      &       0.0006      &       0.2104       \\
			\textbf{Services}      &       0.1366       &       0.0344       &      3.9758     &      0.0002      &       0.0680      &       0.2052       \\
			\hline
		\end{tabular}
	\end{center}
	\caption{Resilience Model PanelOLS Summary}
\end{table}The PanelOLS regression results show strong correlations between resilience and temperature, industrial activity, and service sectors. Temperature has a significant positive impact on resilience, as warmer climates, coupled with appropriate policies, support the implementation of heat-resilient crops and irrigation systems. Precipitation shows a weaker impact, suggesting that rainfall variability must be managed with effective water systems to enhance resilience. Industrial and service sectors contribute significantly to resilience, providing economic diversification and broader social infrastructure, reducing reliance on the more climate-vulnerable agricultural sector.

\section{Discussion and Limitations}
This study presents a comprehensive framework for analyzing climate resilience across LICs. The findings highlight the importance of economic diversification, effective policy implementation, and resource access in building resilience against climate change. The cross-country framework identifies common patterns and differences in resilience, while the localized mapping technique offers targeted insights for intervention. Unlike previous studies that often focus on single-country analyses, this study integrates diverse datasets across multiple LICs, offering a broader perspective on resilience. The use of advanced econometric models and spatial interpolation techniques provides a more nuanced understanding of sectoral adaptation to climate change. The study relies on sparse and sometimes inconsistent data, which may affect the precision of some estimates. The kriging method, while effective for interpolation, assumes spatial correlation that might not always capture complex regional dynamics. Future research could incorporate more comprehensive data collection and consider additional socio-economic variables to refine the analysis further.

\section{Conclusion}

This study enhances the understanding of climate resilience in LICs by introducing a novel cross-country comparative framework and localized mapping technique. The analysis provides policymakers with tools to identify resilience patterns, prioritize interventions, and allocate resources effectively. By highlighting the role of economic diversification, policy implementation, and resource management, the study offers actionable insights to improve climate adaptation strategies at both national and regional levels. Together, these contributions provide a foundation for more informed and effective climate resilience planning in LICs.

\end{document}